\documentclass[a4paper,11pt]{article}
\usepackage{graphicx}
\usepackage{amsmath, amssymb}
\usepackage{hyperref}
\usepackage{booktabs}
\usepackage{float}
\usepackage[a4paper, margin=1.0 in]{geometry}

\title{Automation and Feature Selection Enhancement with Reinforcement Learning (RL)}
\author{Sumana Sanyasipura Nagaraju\thanks{Portland State University, \href{mailto:sumanasn@pdx.edu}{sumanasn@pdx.edu}}}
\date{}
\begin{document}
\maketitle
\section{Abstract}
Effective feature selection, representation and transformation are principal steps in machine learning to improve prediction accuracy, model generalization and computational efficiency. Reinforcement learning provides a new perspective towards balanced exploration of optimal feature subset using multi-agent\cite{article} and single-agent models. Interactive reinforcement learning integrated with decision tree improves feature knowledge, state representation and selection efficiency, while diversified teaching strategies improve both selection quality and efficiency. The state representation can further be enhanced by scanning features sequentially along with the usage of convolutional auto-encoder\cite{zhao2020simplifyingreinforcedfeatureselection}. Monte Carlo-based reinforced feature selection(MCRFS)\cite{liu2021efficientreinforcedfeatureselection}, a single-agent feature selection method reduces computational burden by incorporating early-stopping and reward-level interactive strategies. A dual-agent RL framework\cite{fan2022featureinstancejointselection} is also introduced that collectively selects features and instances, capturing the interactions between them. This enables the agents to navigate through complex data spaces. To outperform the traditional feature engineering, cascading reinforced agents are used to iteratively improve the feature space, which is a self-optimizing framework\cite{xiao2022selfoptimizingfeaturetransformation}. The blend of reinforcement learning, multi-agent systems, and bandit-based approaches offers exciting paths for studying scalable and interpretable machine learning solutions to handle high-dimensional data and challenging predictive tasks. 

\section{Introduction}
In machine learning, one of the most important challenges lies in selecting the right feature subset for subsequent predictions as it impacts the model performance directly.  Classic feature selection methods like filter methods, wrapper methods and embedded methods are fairly efficient but they also show some limitations. Recent advancements incorporate reinforcement learning to automate feature subspace exploration as this technique can interact with environments, learn from action rewards, balance exploration and exploitation which were not satisfied by traditional methods. Interactive Reinforcement Learning(IRL) has been introduced to reduce the exploration space by involving external trainers to guide the agent to make the right decisions.

The gap between automation and efficiency is narrowed by introducing a diversity-aware interactive reinforcement learning(IRL)\cite{fan2020autofsautomatedfeatureselection} approach where an interactive learning feature is integrated with multi-agent RL to apply the prior knowledge of external trainers. To improve feature selection, structured feature knowledge is fed to the decision tree through a feedback-improvement loop. External trainers are brought to diversify the advice by selecting a set of agents randomly which adds robustness to learn a wide range of knowledge. 

The complexity of the Multi-Agent Reinforcement Learning(MARL)\cite{article} framework increases with the number of agents as each feature is considered as an agent and puts pressure on the computational platform. This problem is solved by designing a single agent which scans through the features one by one to decide the selection. To achieve optimal scanning order, relevance of each feature with the label is calculated. With the dynamically changing feature selection process, the state representation needs to be derived, and this is solved by a convolutional auto-encoder. Another framework with Monte Carlo method is introduced to tackle large datasets where the action space is too large for the agent to explore directly. Two policies are designed, i.e. one behavior policy to generate training data and one target policy to generate the final feature subset where a reward-level interactive strategy and early stopping criteria are established reducing the computational overhead. 

Dual-agent interactive reinforcement learning framework is proposed to jointly select features and instances\cite{fan2022featureinstancejointselection} to interact with each other. Two agents are created to select the optimal feature subset and optimal instance subset. To uncover optimal feature space the original features are clustered into different feature groups. Group-operation-group based generation\cite{wang2022groupwisereinforcementfeaturegeneration} approach is designed to accelerate representation space reconstruction. This allows them to explore feature space faster and gain enough reward to learn effective policies. Decision making is improved by graph-based representation to decouple Q-learning process and for better feature relationships. Feature space transformation is done efficiently by adapting bandit-based models. Multi-armed bandit reformulates the problem into combinatorial multi-armed bandit(CMAB)\cite{liu2021multi} where each feature refers to an arm and a subset of features is selected for each iteration.  The process becomes more fast and scalable because of the balanced exploration and exploitation along with reward functions. 

Further, the model performance is optimized by rewarding meaningful feature interactions and feature space is reconstructed as a structured decision-making process. Overall, the traditional feature selection, transformation and engineering is outperformed by modern adaptive and efficient automated strategies which makes it more scalable and computationally efficient.

\section{Background}
The fundamental component for a wide range of data mining and machine learning tasks is to select and extract the important features to provide better interpretation and explanation. Traditional Feature selection methods include filter methods where the features are ranked by relevance scores and only top-ranking features are selected. They are evaluated based on their statistical characteristics, e.g., Fisher score, correlation coefficient and information gain. Wrapper methods use predetermined classifiers to evaluate the quality of selected features. As part of the classification model, embedded methods such as LARS, LASSO\cite{article1} and decision trees incorporate feature selection in the training phase. 

Beside the feature selection techniques, conventional feature transformation approaches involve feature engineering to extract a transformed representation of data. Despite being able to construct an explainable feature set, traditional approaches face several setbacks such as computational costs, ignoring the feature dependencies and interactions between features. Their inability to adapt to exponentially growing feature subset space make it a NP-hard problem. These techniques are labor-intensive, lack interpretability and limit the ability to automate the extraction process.

To overcome these challenges, reinforcement learning technology has demonstrated significant improvement by tackling the feature selection problem with reinforced feature selection. A framework is designed where the agents interact with each other as well as the environment by taking actions. Therefore, there is continuous learning and adaptation. With the reward scheme designed, it can refine its choices after receiving feedback to improve predictive performance. Early stopping criteria is introduced to prevent unnecessary computations. This provides flexibility to reinforcement learning to optimize feature selection and transformation compared to the conventional approaches in scalability, interpretability and automation.

\section{Reinforcement Learning Methodologies for Feature Selection}
\begin{figure}[h]
    \centering
    \includegraphics[width=1.0\textwidth]{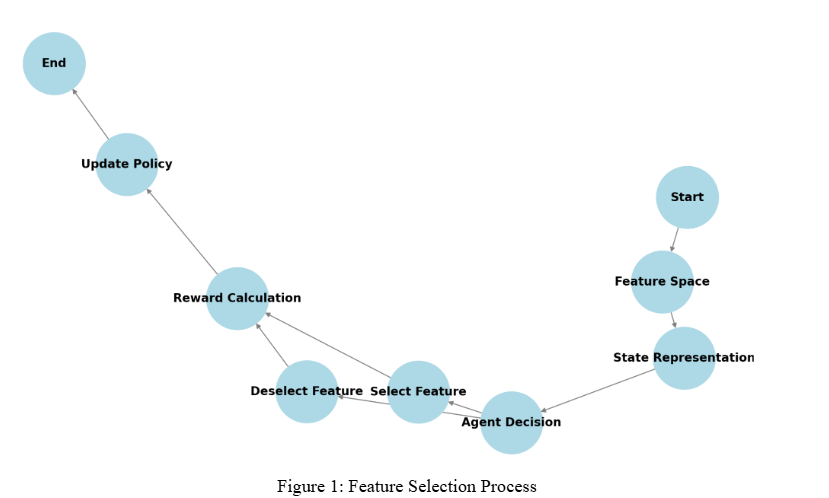}
    \caption{Feature Selection Process}
    \label{fig:feature_selection}
\end{figure}

According to the optimal decision making strategies and automating feature subspace exploration, feature selection and transformation methods can be grouped into different categories. Reinforcement learning has remodeled feature selection by outlining it as a step-by-step decision making problem. Through a reward-scheme approach, the agent learns optimal feature subset by receiving the feedback about their actions. RL-based models are highly efficient and scalable as they can adapt to the environment and provide better state representation of the data based on the data patterns. Below is the generic flow of the feature selection process in reinforcement learning\ref{fig:feature_selection}.

\subsection{Multi-Agent Reinforcement Learning Feature Selection (MARLFS)}
In the Multi-agent reinforcement Learning Feature Selection(MARLFS)\cite{article} framework, each agent is considered as a feature and multiple agents explore the feature spaces simultaneously automating the selection process. It is a MARL task, where agents are independent of each other and make binary decisions whether to select or deselect itself on the basis of learned policies. To maximize the prediction accuracy and lower the redundancy, a reward function is designed where the actions that generate higher accuracy will receive a higher reward.
The framework consists of two stages, i.e., control stage and training stage. In the control stage, each action taken by the feature agent in the current state will decide the next state’s recommended actions based on the network policies. To determine the appropriate state representation, three different methods are developed.
\begin{itemize}
\item Meta Descriptive Statistics for Feature Selection: It extracts the descriptive statistics such as standard deviation, minimum and maximum from the selected data matrix.
\item Static subspace graphs based autoencoder: It is used for learning representations of the reconstructed output which is a fixed-length state vector.
\item Dynamic- Graph Based Convolutional Network(GCN): It creates a feature correlation graph to capture relationships between features.
\end{itemize}
In the training stage, agents use Deep Q-Network\cite{article2} to train their policies and obtain the maximum rewards for long term through experience replay. To guarantee the independence of the selected samples stored in the memory, a Gaussian mixture model(GMM) is suggested which is based on generative rectified sampling algorithm. Here, a set of memory samples is taken as input  where selected actions are clustered into one group and deselected actions are clustered into another one generating high-quality datasets by training selected actions group batchwise with Expectation Maximization(EM)\cite{10.1111/j.2517-6161.1977.tb01600.x} algorithm while replacing low-reward samples each time.

To accelerate the feature space, MARLFS\cite{article} uses two exploration strategies. One is Rank-Based Softmax sampling where the samples are ranked according to the measure of their importance which lowers the computational burden compared to GMM. Another one is through Interactive Reinforcement Learning(IRL) which uses the K-Best Selection method for advising the agents to explore features in a good direction independently avoiding overfitting as it uses pre-trained features.

\subsection{Interactive Reinforced Feature Selection (IRFS)}
IRFS\cite{fan2020interactivereinforcementlearningfeature} framework develops a diversified interactive learning approach by integrating external trainers with decision tree-based feedback to optimize the feature selection in large feature space. The main stages of the framework are 1) interaction with trainers 2) State representation 3) Reward schemes

\textbf{Interaction with Trainers:}
This framework updates the actions based on the advice given by external trainers. The concept of participated/non-participated features is introduced, where only the participated agents are influenced to dynamically change input to trainers. The participated features are further divided into assertive features, where features are selected by the policies, and hesitant features, where features are deselected by the policies. Hesitant features take advised actions from external agents such as KBest and Decision Tree-based trainers. Additionally, a new strategy is employed to increase decision-making efficiency which allows agents to learn from different trainers at different periods, known as the Hybrid Teaching strategy. Each action of the agent is denoted by \( \{a_1, a_2, \dots, a_N\} \). For \( i \in N \), \( a_i = 1 \) means the agent decides to select feature \( i \), while \( a_i = 0 \) means it decides to deselect feature \( i \).

\textbf{State Representation:}
Graph Convolutional Networks (GCN) represent the feature subset using nodes consisting of directed and indirect edges to model the correlation between features. A decision tree is obtained by extracting these relationships. Using the connected feature-feature graph, the decision tree provides feedback, and they are further trained on the selected feature subset to get improved state representation. Two methods are used to represent the state:
\begin{itemize}
    \item Feature importance is highlighted by the weighted sum of the node.
    \item The average sum of the node creates a generalized state vector.
\end{itemize}

\textbf{Reward Scheme:}
The reward function is measured by combining the predictive accuracy (\( Acc \)), feature correlation (\( R \)), and action records. It is given by:
\begin{equation}
    r_{t,i} =
    \begin{cases}
        W^t_i (Acc - \beta R), & \text{if } a^t_i = 1 \\
        0, & \text{if } a^t_i = 0
    \end{cases}
\end{equation}
where \( W^t_i \) represents a weight derived from historical selection frequency, \( \beta \) is a tuning parameter, \( Acc \) is the predictive accuracy, and \( R \) represents the correlation among selected features.\\
Two personalized reward schemes are proposed to improve the way to calculate reward.
\begin{itemize}
\item Decision Tree Structured Feedback - Feature importance from the decision tree feedback influences the reward.
\item Historical Action records - Frequency ratio of the corresponding agents decides the reward towards the agent with action records considered.
\end{itemize}

\subsection{Diversity-Aware Interactive Reinforced Feature Selection}
This framework is an extension of IRFS\cite{fan2020interactivereinforcementlearningfeature} which includes diversification of teaching along with external trainers improving feature selection efficiency. At different stages agents are guided by different trainers. From step 0 to step T and from step T to 2T are offered advice by different trainers at different steps \cite{fan2020autofsautomatedfeatureselection}. Agents explore and self learn the features without any trainer. This achieves better accuracy and computational efficiency.

\subsection{Scanning-Based Single-Agent Reinforcement Learning}
Single-agent RL models are employed to handle the task of selecting multiple features by sequentially scanning them based on prior scores. This can handle even high-dimensional datasets successfully. 
\begin{itemize}
 \item A scanning strategy is developed based on the feature relevance to decide the order of scanning. Highly relevant features are considered first after calculating the relevance between features and target class, called information gain(IG)\cite{zhao2020simplifyingreinforcedfeatureselection}.  
 \item To assess the performance in automatic feature selection, the reward function combines both predictive accuracy and feature redundancy. A predefined classifier is used to calculate the predictive accuracy leading the scanning agent to receive positive feedback. For feature redundancy, Pearson correlation coefficient is used to measure the duplicate information between the features by giving negative feedback.
 \item Reinforcement learning needs a fixed-length vector for state representation because the selected feature subset keeps updating. Therefore, CNN is integrated with auto-encoder so that it concatenates the spatial bins into a layer to generate the fixed-length vector and is known as convolutional auto-encoder(CAE). To maintain the size of the input image and output image a decoder is constructed with an inverse spatial pyramid pooling layer\cite{He_2014}.
 \item Deep Q-network(DQN)\cite{article2} is used for training where past experiences are stored in a memory and used to update Q-values with a $\epsilon$-greedy approach while the single agent scans through the network with a tracker to point to the feature progression(one-hot encoder). This reduces computational complexity and automates feature selection balancing exploration and exploitation of feature space. 
\end{itemize}

\subsection{Monte Carlo-Based Reinforced Feature Selection (MCRFS)}
The Monte Carlo-based Reinforcement Feature Selection (MCRFS) technique employs a single-agent approach to traverse the entire feature set sequentially to select/deselect the feature. Markov Decision Process (MDP) is modeled for feature selection in the form of state-action pair where the state is represented by a feature subset and each action is represented by each feature decision.

The agent is guided by the Monte Carlo method where a behavior policy is used to decide the feature selection episodes.\cite{liu2021efficientreinforcedfeatureselection} The episodes evaluate the target policy and optimize as it is responsible for taking actions in the next state. This is done using the Bellman equation:
\begin{equation}
    \pi^*(s) = \arg\max_a Q^\pi(s,a)
\end{equation}
where \( \pi \) is the target policy, and \( (s,a) \) represent state-action pairs. Because the policies are different, the reward is adjusted by applying importance sampling weight, reducing variance in training. The weights are updated incrementally instead of calculating the sampling weights for every decision. If skew samples are generated, i.e., sample weights with low importance, the agent stops the traversal. This strategy is known as the Early Stopping strategy\cite{liu2021efficientreinforcedfeatureselection}, which prevents unnecessary exploration of feature subsets, reducing computational overhead. The stopping probability is given by:
\begin{equation}
    p_t = \max(0,1 - \frac{p_t}{v})
\end{equation}
where \( p_t \) is the importance sampling weight and \( v \) is the stopping threshold.

Training speed is further accelerated by introducing the Interactive Reinforcement Learning framework\cite{fan2020interactivereinforcementlearningfeature}, where an external advisor provides feedback based on the benefits driven by the action. The reward function is modified to combine both environment feedback and external utility guidance to help the agent explore the optimal policy efficiently.

\subsection{Dual-Agent Interactive Reinforced Selection (DAIRS)}
A dual agent RL is introduced as a variant of multi-agent RL to perform two different tasks. One is a feature agent whose concentration is on identifying feature-feature correlation to get an optimized feature subset while the other agent is an Instance agent modelling the instance-instance correlation enabling the interaction between each feature and each agent\cite{fan2022featureinstancejointselection}. This optimization improves the efficiency of agent learning.

Feature agent scans the actions sequentially selecting/deselecting the feature at each step. Instance agent also follow a similar method. To globally optimize the dataspace, the dual agents collaborate with each other forming a sub-matrix of the input data. This collaborative-changing environment allows both the agents to interact with each other improving the model performance. To solve the dynamically changing dimension problem, DAIRS\cite{fan2022featureinstancejointselection} introduced an image processing technique where the sub-matrix is represented as a 2D image and applies convolution operation to generate fixed-length state representation. This framework also uses Deep Q- Network\cite{article2} to train its policies with experience replay. Additionally, it uses IRL integrated with external trainers that gives advisory feedback. Random Forest Trainer is employed\cite{fan2022featureinstancejointselection} to measure the feature importance to prioritize the feature by learning a set of decision trees. Likewise,  an Isolation Forest Trainer\cite{fan2022featureinstancejointselection} is used to filter out noisy instances enhancing the training efficiency.

\subsection{Group-Wise Reinforcement Feature Generation (GRFG)}
Group-wise Reinforcement Feature Generation\cite{wang2022groupwisereinforcementfeaturegeneration} iteratively reconstructs the feature space using the feature-crossing method by nesting feature generation and selection, producing better feature groups which improves explainability of representation space.

To begin with, features are clustered into different feature groups to maximize inter-group feature similarity while minimizing intra-group distinctness by developing a distance measure to check distance between the two groups. The distance between the nearest feature groups can be used as stopping criteria to validate the difference in information between feature groups. A cascading agent-based reinforcement learning method\cite{wang2022groupwisereinforcementfeaturegeneration} is adapted, where the three agents are following sequential decision-making. The first agent selects the first feature group, and the second agent picks an operation depending on the first agent’s selection. The third agent selects the second feature group based on the selections made by the first and second agent. Here, each decision will affect the subsequent decisions and make sure that feature transformations are optimal.

There are two generation scenarios as an outcome of selection results that can either be a unary operation or a binary operation which is based on the relevant information with respect to the target variable. K new features are generated by selecting top-K dissimilar features by crossing both the groups. This is fed into downstream ML task, where the agents update their policies after receiving feedback about their predictive performance. A feature selection step is employed at each iteration to avoid excessive feature expansion.

\subsection{ Self-Optimizing Feature Transformation}
The Self-Optimizing Feature Transformation framework\cite{xiao2022selfoptimizingfeaturetransformation} is an extension of GRFG. The main aim here is to improve the state representation through graph-based embeddings and to mitigate Q-value overestimation\cite{vanhasselt2015deepreinforcementlearningdouble}.

The framework constructs a feature-feature-graph where the graph is converted into graph embedding using message-passing mechanisms preserving both feature set knowledge and structural dependencies. To handle Q-value overestimation\cite{vanhasselt2015deepreinforcementlearningdouble}, the framework uses a dual Q-network approach, where the actions are selected by one network and Q-values are estimated by the other. Since the networks are independent of each other, the parameters are updated asynchronously which reduces Q-values of the networks connected to the same action.

\subsection{Combinatorial Multi-Armed Bandit-Based Feature Selection (CMAB-FS)}
The combinatorial multi-armed bandit(CMAB)\cite{liu2021multi} framework refers to each feature as an arm where a super arm is generated which decides the selected feature subset in an iterative manner increasing the accuracy while minimizing redundancy. It can achieve parallel performance with lower computational cost when compared to multi-agent reinforcement learning.
Selection methods employed by CMAB-FS are as follows:
\begin{itemize}
\item Combinatorial Multi-Armed Bandit Generative Feature Selection (CMAB-GFS)\cite{liu2021multi}: A generative model is developed to construct the super arm. Each arm is assigned a distribution to generate the probability of that arm being a super arm with a value-based ranking. To fasten the exploration, $\epsilon$-greedy exploration strategy is applied which brings randomness to improve generalization of suboptimal subsets.
\item Combinatorial Upper Confidence Bounds-Based Feature Selection (CUCB-FS)\cite{liu2021multi}: In this model, ranking-based strategy is applied to form a super arm by selecting the top K arms whose confidence scores are predetermined according to the historical records. The Upper Confidence Bound (UCB) approach ranks features with high uncertainty, ensuring an optimal trade-off between exploration and exploitation.
\end{itemize}

\subsection{Hierarchical Reinforced Feature Space Reconstruction}
HRecon\cite{azim2024featureinteractionawareautomated} models feature space reconstruction as a nested reinforcement learning process, where agents generate new features and control the feature space size simultaneously. It consists of three learning agents. An operation agent to regulate categorical features to fit in. Two feature agents to efficiently explore the feature space and measure the effectiveness of the feature interaction in the subsequent steps or downstream tasks. 

Rewards are determined based on 1) operation validity of transformations matching the feature types. 2) Feature interaction strength which quantifies the feature interaction using H-statistics. 3) Downstream task performance which evaluates classification/regression accuracy. This reward-augmented strategy improves learning efficiency with systematic exploration. It optimizes the reconstruction feature space by dynamically adjusting its size using the feature selection mechanism. When the feature set is very large, K-best selection only selects the top informative features.

\section{Experimental Findings}
Below is the table of comparison of results found after conducting the experiment for the various methodologies.

\begin{table}[H]
    \centering
    \renewcommand{\arraystretch}{1.5} 
    
    \begin{tabular}{|p{3cm}|p{3cm}|p{3cm}|p{3cm}|p{3cm}|}
        \hline
        \textbf{Methodology} & \textbf{Metrics of Evaluation} & \textbf{Accuracy} & \textbf{Computational Cost} & \textbf{Feature Selection Quality} \\
        \hline
        Multi-Agent Reinforcement Learning Feature Selection\cite{article} & Accuracy, Feature Redundancy Reduction & 87.0 (RF), 84.0 (LASSO) & Execution time: 1.44s per step (MARLFS) & Lower redundancy, better subset accuracy \\
        \hline
        Interactive Reinforced Feature Selection\cite{fan2020interactivereinforcementlearningfeature} & Exploration Efficiency, Best Accuracy &  approx. 93.0(Spambase), 91.0(ICB), 98.0(MUSK), 79.0(FC) & Significantly improved exploration efficiency & Stable selection across datasets, outperforming baselines \\
        \hline
        Diversity-Aware Interactive Reinforced Feature Selection\cite{fan2020autofsautomatedfeatureselection} & Accuracy, Stability Across Datasets & 93.0(Spambase), 91.0(ICB), 98.2(MUSK), 81.0(FC) & Stable selection with minimal computational overhead & More consistent feature sets vs RFE, mRMR \\
        \hline
        Scanning-Based Single-Agent Reinforcement Learning\cite{zhao2020simplifyingreinforcedfeatureselection} & Computational Efficiency, Feature Selection Quality & Higher efficiency vs traditional methods & Lower cost compared to wrapper methods & Compact feature selection improves ML performance \\
        \hline
        Monte Carlo-Based Reinforced Feature Selection\cite{liu2021efficientreinforcedfeatureselection} & Predictive Accuracy, Redundancy Minimization & Improved accuracy over LASSO, RFE & Reduced search space via early stopping & Optimal trade-off between relevance and redundancy \\
        \hline
        Dual-Agent Interactive Reinforced Selections\cite{fan2022featureinstancejointselection} & Dual-Agent Optimization, Training Speed & Consistent accuracy gain with dual-agent learning & Interactive RL accelerates convergence & Feature-instance co-optimization improves stability \\
        \hline
        Group-wise Reinforcement Feature Generation\cite{wang2022groupwisereinforcementfeaturegeneration} & Feature Generation Quality, Interpretability & Superior feature transformation in structured datasets & Cascading selection reduces computation & Generated features maintain interpretability, reduce noise \\
        \hline
        Self-Optimizing Feature Transformation\cite{xiao2022selfoptimizingfeaturetransformation} & Graph-Based Representation, Q-Value Correction & Enhanced accuracy with graph-based representation & Efficient Q-value correction enhances learning & Graph-based feature selection preserves structure \\
        \hline
        Combinatorial Multi-Armed Bandit-Based Feature Selection\cite{liu2021multi} & Exploration-Exploitation Balance, Feature Ranking & Better trade-off in accuracy vs computational cost & Parallel selection minimizes redundancy & Multi-armed bandit finds optimal subsets faster \\
        \hline
        Hierarchical Reinforced Feature Space Reconstruction\cite{azim2024featureinteractionawareautomated} & Hierarchical RL, Feature Interaction Strength & Optimized hierarchical learning for scalable selection & Hierarchical structure reduces complexity in large datasets & Hierarchical feature adaptation refines selection iteratively \\
        \hline
    \end{tabular}
    \caption{Experimental results of various RL methodologies}
    \label{tab:comparison}
\end{table}

\section{Applications of Reinforcement Learning}
Reinforcement Learning has shown significant success in optimal feature selection in diverse real-world areas. RL can strengthen decision-making by selecting relevant features in various environments which includes fields like medical diagnostics, finance, Robotics etc.,.

\textbf{Medical Diagnostics:} RL is used in Sepsis Treatment where it learns the optimal policies by clustering patient states and optimizes treatment steps for administering fluids  and vasopressors to increase the 90-day survival rates. It is also used in Medical Image Report Generation where CNN-based image coding is applied to automate radiology reports\cite{li2019reinforcementlearningapplications}.

\textbf{Finance:} By improving pricing and Monte Carlo Methods with RL based simulations, American options pricing is enhanced. Stock trading decisions are modeled with bid-ask spread with the help of RL\cite{li2019reinforcementlearningapplications}.

\textbf{Robotics:} Dexterous Robotic Manipulation employs RL to handle the robot hand control for precise object manipulation avoiding the use of manually designed controllers. RL is also incorporated in Legged Robots to achieve efficient locomotion\cite{li2019reinforcementlearningapplications}.

\section{Future Challenges and Research Directions}

Automation and optimization are the two major benefits offered by Reinforcement learning (RL)-based feature selection but faces challenges associated with scalability, interpretability, adaptability, efficiency, and fairness. High-dimensional feature spaces increase computational complexity, making multi-agent RL frameworks expensive and difficult to deploy in real-time applications. It takes considerably more time for Monte Carlo-based to select the optimal features, which adds the need to research more hierarchical RL methods and graph-based embeddings to improve scalability. Redundancy and interpretability take a backseat to accuracy in the existing RL reward schemes causing poor feature selection. Hybrid reward functions incorporating accuracy, redundancy reduction, and feature importance could enhance decision-making.

Interpretability remains a significant challenge, as RL models often function as black boxes with limited transparency. Another concerning element is adaptability because most of the RL-based feature selection models require retraining for different datasets. Future work should focus on transfer learning and continual learning approaches to enable RL models to dynamically adapt to changing data distributions. RL also faces challenges with multi-modal data as it is important to develop specialized architectures which can handle structured and unstructured features effectively.

RL models require significant computational resources which brings a major concern of energy efficiency. Optimizing RL for cloud and edge environments through energy-efficient algorithms and cost-aware training strategies could reduce operational costs. RL-based features can introduce biases in applications that are very impactful like finance and healthcare. Therefore, it is necessary to tackle issues related to privacy and fairness. Future research should explore fairness-aware RL frameworks and privacy-preserving approaches such as federated learning to ensure ethical feature selection. Addressing the above said challenges will strengthen RL-based feature selection's scalability, interpretability, adaptability, and ethical considerations, making it a more reliable tool for real-world machine learning applications.

\section{Conclusion}
Machine learning models are revolutionized  by techniques like feature selection and transformation to manage high-dimensional data. By collaborating with advanced RL models like multi-agent models, and hybrid optimization frameworks, these methodologies bring in opportunities for more adaptive, scalable, and interpretable machine learning pipelines. These methodologies select the features with respect to the environment which keeps updating resulting in the improved predictive accuracy along with the enhancement in computational efficiency. Future developments should concentrate on improving the computational efficiency and feature interpretability so that the techniques of reinforcement learning can be incorporated to applications in a variety of fields, including autonomous systems, finance, and medical.

\bibliographystyle{IEEEtran} 
\bibliography{main}
\end{document}